\documentclass[10pt,twocolumn,letterpaper]{article}

\usepackage[pagenumbers]{cvpr} 

\usepackage[accsupp]{axessibility}

\usepackage{graphicx}
\usepackage{amsmath}
\usepackage{amssymb}
\usepackage{booktabs}
\usepackage{siunitx}
\usepackage{multirow}

\usepackage{soul}  

\usepackage[pagebackref,breaklinks,colorlinks]{hyperref}

\usepackage[capitalize]{cleveref}
\crefname{section}{Sec.}{Secs.}
\Crefname{section}{Section}{Sections}
\Crefname{table}{Table}{Tables}
\crefname{table}{Tab.}{Tabs.}

\newenvironment{para_packed_item}{
\begin{changemargin}{-1.75em}{0em} 
\begin{itemize}
  \setlength{\itemsep}{1pt}
  \setlength{\parskip}{0pt}
  \setlength{\parsep}{0pt}
   \setlength{\itemindent}{0.9em}
}{\end{itemize}
\end{changemargin}
}

\def\cvprPaperID{1} 
\def\confName{CVPR}
\def\confYear{2023}

\begin{document}

\title{Exploring the Utility of Self-Supervised Pretraining Strategies for the Detection of Absent Lung Sliding in M-Mode Lung Ultrasound}

\author{Blake VanBerlo\textsuperscript{1,3\,*}, Brian Li\textsuperscript{1,3}, Alexander Wong\textsuperscript{1}, Jesse Hoey\textsuperscript{1}, Robert Arntfield\textsuperscript{2,3} \\
\textsuperscript{1}University of Waterloo, \textsuperscript{2}Western University, \textsuperscript{3}Deep Breathe\\
{\tt\small bvanberl@uwaterloo.ca}}

\maketitle

\begin{abstract}

Self-supervised pretraining has been observed to improve performance in supervised learning tasks in medical imaging. This study investigates the utility of self-supervised pretraining prior to conducting supervised fine-tuning for the downstream task of lung sliding classification in M-mode lung ultrasound images. We propose a novel pairwise relationship that couples M-mode images constructed from the same B-mode image and investigate the utility of data augmentation procedure specific to M-mode lung ultrasound. The results indicate that self-supervised pretraining yields better performance than full supervision, most notably for feature extractors not initialized with ImageNet-pretrained weights. Moreover, we observe that including a vast volume of unlabelled data results in improved performance on external validation datasets, underscoring the value of self-supervision for improving generalizability in automatic ultrasound interpretation. To the authors' best knowledge, this study is the first to characterize the influence of self-supervised pretraining for M-mode ultrasound.
\footnote{© 2023 IEEE. Personal use of this material is permitted. Permission from IEEE must be obtained for all other uses, in any current or future media, including reprinting/republishing this material for advertising or promotional purposes, creating new collective works, for resale or redistribution to servers or lists, or reuse of any copyrighted component of this work in other works.}  

\end{abstract}

\section{Introduction}
\label{sec:intro}

Pneumothorax (PTX) is a potentially life-threatening acute condition in which air occupies the space between the pleura of the lungs, resulting in collapse of the lung. Rapid identification of PTX is crucial in emergency, critical, and acute care settings to expedite medical intervention. Point-of-care lung ultrasound (LUS) is a quick, inexpensive, portable, imaging examination that does not expose patients to radiation. Despite its low prevalence compared to chest radiographs, LUS has been shown to exhibit superior diagnostic performance for the diagnosis of PTX~\cite{Nagarsheth2011, Alrajhi2012}. The lung sliding artefact, caused by the normal motion of the pleura, has been described as a means to rule out PTX~\cite{Lichtenstein1995}. Notably, the presence of lung sliding excludes a diagnosis of PTX within the purview of the ultrasound probe~\cite{Lichtenstein1995}. Conversely, PTX is likely present when lung sliding is absent.

Previous studies have demonstrated that deep convolutional neural networks (CNN) can be trained to distinguish between the presence and absence of lung sliding on motion mode (M-mode) ultrasound images~\cite{Javsvcur2021, VanBerlo2022}. Prior studies were limited by the amount of labelled data available for training and evaluation. Furthermore, previous studies initialized their networks using weights pretrained on the ImageNet dataset~\cite{Deng2009}. Despite the fact that M-mode images are profoundly distinct from the natural images present in ImageNet, it is common for medical imaging studies to leverage ImageNet-pretrained weights. They are publicly available for several common architectures and are able to extract low-level features present in medical images. Unfortunately, there are no publicly available equivalents for M-mode images, let alone LUS.

Self-supervised learning (SSL) is a representation learning strategy applicable in the absence of labelled data. CNNs pretrained using SSL have exhibited superior performance and label efficiency compared to fully supervised counterparts~\cite{Chen2020,Grill2020,Zbontar2021,Bardes2022}. Broadly, SSL pretraining consists of training a deep neural network to solve a \textit{pretext task}, whose solution can be computed from unlabelled examples. The pretrained weights may be fine-tuned to solve a \textit{downstream task} for which labels are present. This study explores the impact of self-supervised pretraining for the downstream task of detecting absent lung sliding in M-mode LUS, varying the choice of SSL method, weight initialization, data augmentation strategy, and inclusion of unlabelled data. Crucially, we demonstrate that incorporating large volumes of unlabelled M-mode images during the pretraining phase improves the performance of a fine-tuned classifier on external datasets. More specifically, our major contributions are as follows:

\begin{para_packed_item}
    \item A pairwise relationship for contrastive and non-contrastive learning that is specific to M-mode images
    \item A data augmentation pipeline specific to M-mode images in the context of pretraining
    \item A comprehensive investigation of factors that influence the utility of SSL pretraining for the downstream task of absent lung sliding detection, such as label efficiency, ImageNet initialization, and data augmentation
    \item Evidence that the inclusion of unlabelled data results in improved generalization to external datasets for absent lung sliding detection
\end{para_packed_item}

\begin{figure*}[ht]
    \centering
    \includegraphics[width=\linewidth]{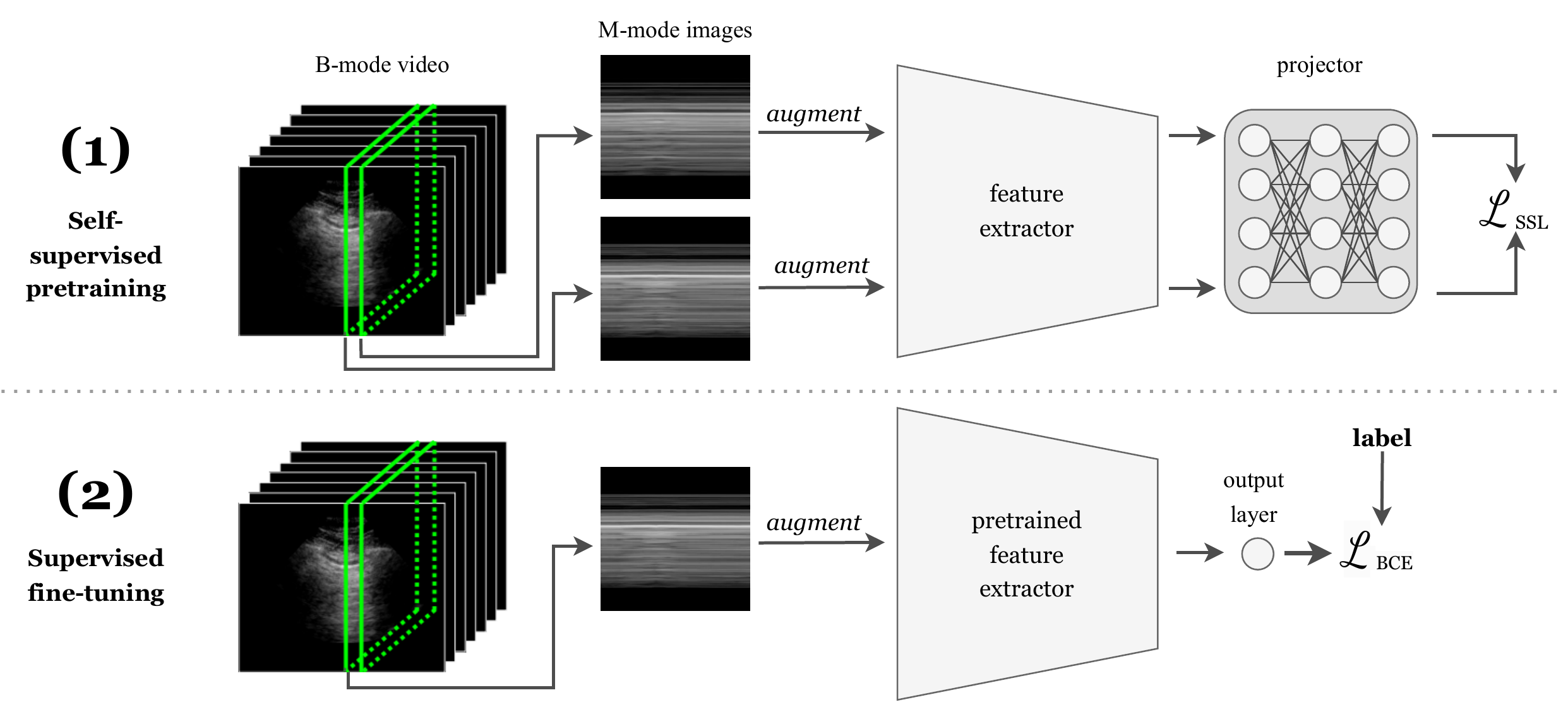}
    \caption{An overview of the methods in this paper. (1) Self-supervised pretraining using pairs of M-mode images extracted from the same B-mode LUS video. Both M-mode images are passed through a joint-embedding SSL architecture, consisting of a CNN feature extractor and multilayer perceptron projector in series. The objective, $\mathcal{L}_{SSL}$, is computed using the embeddings outputted by the projector. The pretrained CNN feature extractor is retained and the multilayer perceptron projector is discarded. (2) A single-node output layer is appended to the pre-trained feature extractor. The resulting model is fine-tuned to solve the downstream task of absent lung sliding detection by minimizing the binary cross entropy loss with respect to the labelled training set.}
    \label{fig:overview}
\end{figure*}

\cref{fig:overview} summarizes our methods. To the best of our knowledge, no study has investigated the efficacy of SSL for M-mode ultrasound tasks.

\section{Related Work}
\label{sec:related-works}

\subsection{Lung Sliding Classification}
\label{subsec:lung-sliding-classification}

Multiple studies have explored the use of CNNs for automatically identifying absent lung sliding in LUS M-modes. Jaščur \etal~\cite{Javsvcur2021} were the first to attempt this task. Using a dataset of $48$ videos acquired from $48$ post-thoracic surgery patients with a single ultrasound probe, their model achieved a sensitivity of $0.82$ and specificity of $0.92$. VanBerlo \etal~\cite{VanBerlo2022} developed a binary classifier using a dataset of $2535$ examples acquired from hundreds of patients with variable probes. When evaluated on a test set of $540$ examples, their model attained a sensitivity and specificity of $0.935$ and $0.873$ respectively.

\subsection{Self-supervised Learning in Computer Vision}

Self-supervised learning methods in computer vision can be categorized in various manners, based on the pretext task. Generative methods, such as image colourization~\cite{Zhang2016} and inpainting~\cite{Pathak2016}, often involve reconstructing a corrupted image. Predictive tasks, on the other hand, consist of learning to predict or undo a transformation applied to an image. For example, the jigsaw pretext task is characterized by unscrambling randomly permuted rectangular patches of an image~\cite{Noroozi2016}.

Several contemporary approaches adopt the joint-embedding architecture, in which representations of paired samples are compared. In a contrastive learning task, objectives are designed to minimize the distance between representations of paired examples (i.e., \textit{positive pairs}) and maximize the distance between those of examples from different pairs (i.e., \textit{negative pairs})~\cite{He2020, Chen2020}. Non-contrastive methods aim to minimize the difference between positive pairs, disregarding negative pairs~\cite{Grill2020, Chen2021, Zbontar2021,Bardes2022}. Recent methods have added regularizers to mitigate a degenerate solution in non-contrastive learning in which representations for all examples trend toward zero vectors~\cite{Zbontar2021, Bardes2022}.

In the context of joint-embedding methods, the \textit{pairwise relationship} enforces constraints between examples that qualify them as a positive pair. Typically, a positive pair consists of two perturbations of a single example, where each perturbation is sampled from a distribution over image transformations. In their paper demonstrating how SimCLR improves performance and label efficiency in chest X-ray and dermatological image classification, Azizi \etal~\cite{Azizi2021} suggested an alternative, situation-specific pairwise relationship that reflects a stronger inductive bias in the pretrained network -- they considered any two distinct images of the same pathology to be a positive pair. For example, posteroanterior and lateral chest X-rays from the same patient encounter are a positive pair. A recent theoretical analysis by Balestriero and LeCun lends credence to the idea of context-specific positive pairs, providing justification that pretraining using SimCLR~\cite{Chen2020}, Barlow Twins~\cite{Zbontar2021}, or VICReg~\cite{Bardes2022} will improve performance on a downstream supervised learning task, as long as the pairwise relationship aligns with the labels for that task~\cite{Balestriero2022}. The authors' results underline the importance of medical knowledge in designing pretext tasks, motivating the definition of the M-mode pairwise relationship presented in this study.

\subsection{SSL for Medical Ultrasound}

SSL for medical ultrasound is underexplored compared to other medical image modalities, but some studies have investigated its utility for brightness mode (B-mode) ultrasound images and videos. Jiao \etal~\cite{Jiao2020} observed an improvement in performance on the downstream task of fetal plane detection after pretraining a CNN to both reorder the images of a shuffled fetal ultrasound video and predict a transformation that was applied to it. Basu \etal~\cite{Basu2022} explored the benefit of defining negative pairs within the same ultrasound video in addition to those across videos, constructing a contrastive learning procedure in which intra-video pairs are introduced to the model after inter-video pairs. Self-supervised pretraining has also been effective for multiple echocardiography tasks, including atrial fibrillation detection~\cite{Dezaki2021}, left ventricle segmentation~\cite{Saeed2022}, and view identification~\cite{Anand2022}.

\section{Methods}
\label{sec:methods}

\subsection{Dataset}

\begin{figure}[t]
    \centering
    \includegraphics[width=\linewidth]{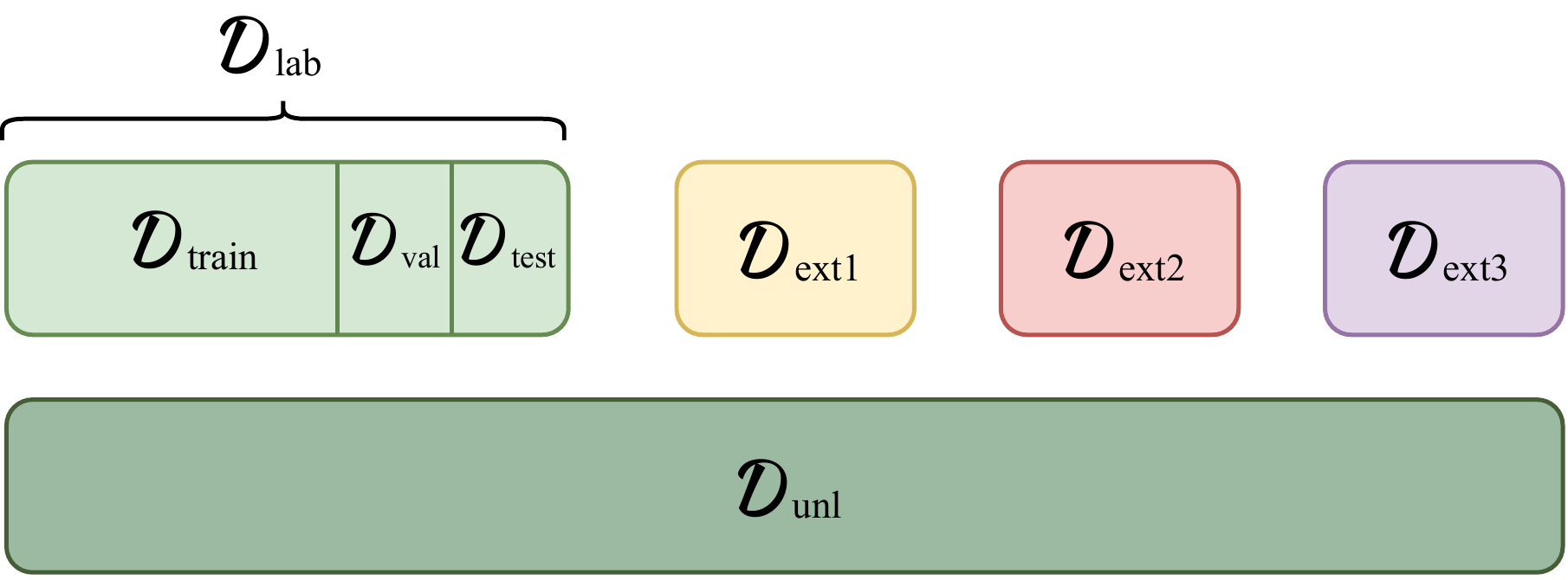}
    \caption{The datasets used in this study. $\mathcal{D}_{\text{lab}}$ is labelled for the downstream lung sliding classification task. It is split randomly by $70\%/15\%/15\%$ into $\mathcal{D}_{\text{train}}$, $\mathcal{D}_{\text{val}}$, and $\mathcal{D}_{\text{test}}$, such that patients do not appear in multiple partitions.}
    \label{fig:dataset-split}
\end{figure}

The datasets used for all training experiments in this study originated from a large, private B-mode LUS database collected from two hospitals within an academic healthcare institution, the use of which is permitted by ethics approval granted by Western University (REB 116838) A portion of this database was previously labelled for the presence or absence of lung sliding by a critical care physician possessing expertise in LUS (hereafter referred to as the \textit{LUS expert}). All LUS videos were divided into \SI{3}{s} segments, with excess frames discarded. The resulting dataset, hereafter referred to as $\mathcal{D}_{\text{lab}}$, contained $4793$ videos. $\mathcal{D}_{\text{lab}}$ was then randomly split by patient into three partitions: a training set of $3254$ videos ($\mathcal{D}_{\text{train}}$), validation set of $743$ videos ($\mathcal{D}_{\text{val}}$), and test set of $796$ videos ($\mathcal{D}_{\text{test}}$). The database was queried for additional videos containing the A-line artefact but that were missing labels for lung sliding. This additional tranche of $14249$ unlabelled videos is referred to as $\mathcal{D}_{\text{unl}}$. \cref{fig:dataset-split} illustrates the dataset split.

To investigate model generalizability, we evaluate on three additional datasets from external healthcare institutions labelled for lung sliding: $\mathcal{D}_{\text{ext1}}$, $\mathcal{D}_{\text{ext2}}$, and $\mathcal{D}_{\text{ext3}}$. \cref{tab:dataset-breakdown} provides details on the label and patient decomposition of all labelled datasets.

\begin{table*}
    \centering
    \begin{tabular}{ccccccccc}
        \toprule
         &  & $\mathcal{D}_{\text{train}}$ & $\mathcal{D}_{\text{val}}$ & $\mathcal{D}_{\text{test}}$ &  $\mathcal{D}_{\text{unl}}$ & $\mathcal{D}_{\text{ext1}}$ & $\mathcal{D}_{\text{ext2}}$ & $\mathcal{D}_{\text{ext3}}$ \\
        \midrule
        \multirow{2}{*}{{\large Lung sliding Present}} & Videos & $2509$ & $564$ & $607$ & - & $84$ & $121$ & $424$  \\
        & Patients & $366$ & $76$ & $72$ & - & $25$ & $55$ & $155$  \\
        \multirow{2}{*}{{\large Lung sliding Absent}} & Videos & $745$ & $179$ & $189$ & - & $26$ & $30$ &  $77$ \\
        & Patients & $199$ & $49$ & $48$ & - & $4$ & $11$ & $40$ \\
        \multirow{2}{*}{{\large Total}} & Videos & $3254$ & $743$ & $796$ & \num{14249} & $88$ & $151$ & $501$  \\
        & Patients & $565$ & $125$ & $120$ & $3113$ & $29$ & $66$ & $195$  \\
        \bottomrule
    \end{tabular}
    \caption{Class decomposition }
    \label{tab:dataset-breakdown}
\end{table*}

\subsubsection{M-modes from B-modes}
\label{subsubsec:mmodes-from-bmodes}

We follow a similar method outlined in~\cite{VanBerlo2022} to extract M-mode images from greyscale B-mode videos. For the binary classification task, the vertical slice of the B-mode video with the maximum total pixel intensity is selected from all possible vertical slices within the horizontal bounds of the pleural line. This heuristic is consistent with the clinical notion that, in the vast majority of cases, the pleural line is the brightest artefact in a LUS image of the upper and middle lobes of the lung. All M-mode images were resized to $128 \times 128$ pixels prior to pretraining.
When pretraining, we retain the top $50\%$ of each video's M-modes, ordered by total pixel intensity. As will be discussed in \cref{subsubsec:pairwise-relationship}, the pairwise relationship for M-modes requires distinct M-mode images from the same video. 

\subsection{Self-Supervised Learning}

\subsubsection{Pretext Tasks}

We investigate three commonly employed joint-embedding self-supervised pretraining techniques from the literature. SimCLR~\cite{Chen2020} is a contrastive learning method that employs a temperature-scaled cross entropy objective. We set the temperature to $\tau = 0.1$. Barlow Twins~\cite{Zbontar2021} is a non-contrastive learning method that minimizes distance between embeddings of pairs and includes an embedding decorrelation regularizer. We employ Barlow Twins with $\lambda = 0.005$ for the weight of the decorrelation regularizer. Inspired in part by Barlow Twins, VICReg~\cite{Bardes2022} is a non-contrastive method that includes a regularizer that minimizes variance across the embedding dimension. We conduct pretraining trials with each of these methods. We set the weights of VICReg's three objective components to $\lambda=25$, $\mu=25$, and $\nu=1$. 

We pretrain for $100$ epochs and use a batch size of $128$ for all experiments. As in~\cite{Azizi2021}, we also investigate the effect of initializing the feature extractors with ImageNet-pretrained weights prior to self-supervised pretraining. After all pretraining experiments, weights of the projector are discarded and the feature extractor are preserved for initialization in the downstream task. We adopt the EfficientNetB0~\cite{Tan2019} architecture as the feature extractor for all experiments, discarding the final block to reduce model capacity. In each experiment, the projector is a multilayer perceptron with $3$ layers of $128$ nodes, with the rectified linear unit activation applied to each hidden layer.

\subsubsection{Pairwise Relationship}
\label{subsubsec:pairwise-relationship}

\begin{figure*}
  \centering
  \begin{subfigure}{0.24\linewidth}
    \includegraphics[width=\linewidth]{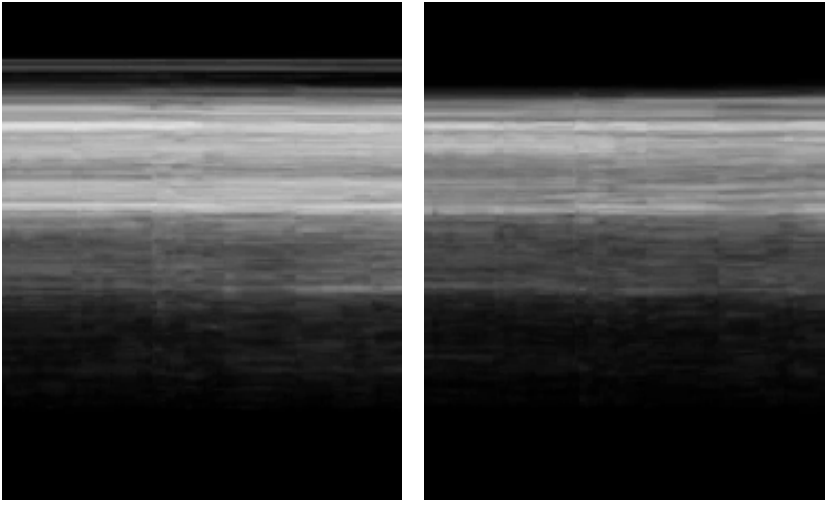}
    \caption{Lung sliding, seashore sign.}
    \label{subfig:sliding1}
  \end{subfigure}
  \hfill
  \begin{subfigure}{0.24\linewidth}
    \includegraphics[width=\linewidth]{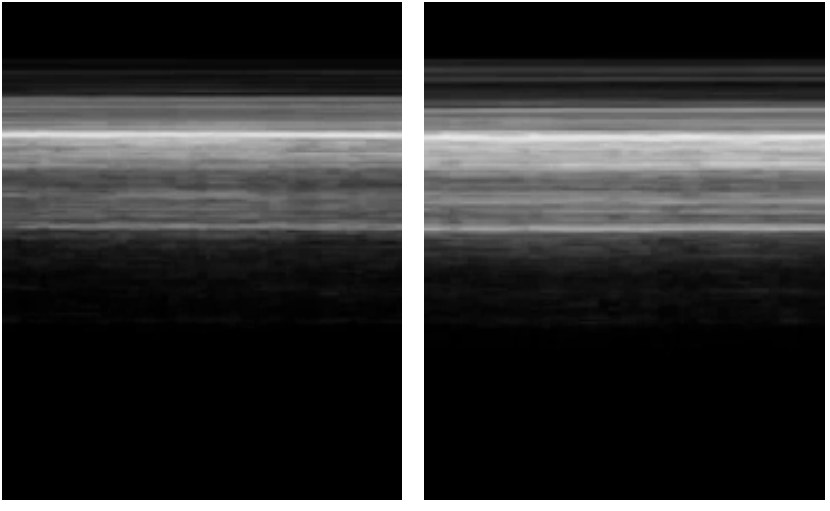}
    \caption{Lung sliding, seashore sign.}
    \label{subfig:sliding2}
  \end{subfigure}
  \hfill
  \begin{subfigure}{0.24\linewidth}
    \includegraphics[width=\linewidth]{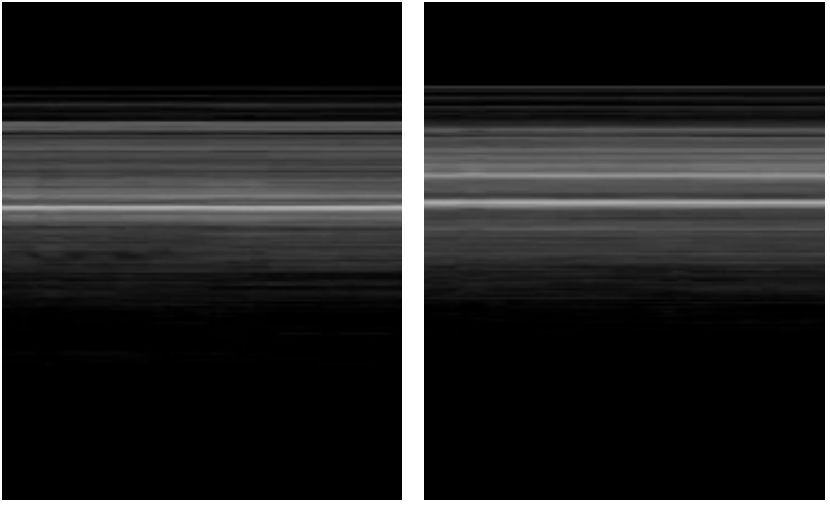}
    \caption{Absent lung sliding, barcode sign.}
    \label{subfig:no-sliding1}
  \end{subfigure}
  \hfill
  \begin{subfigure}{0.24\linewidth}
    \includegraphics[width=\linewidth]{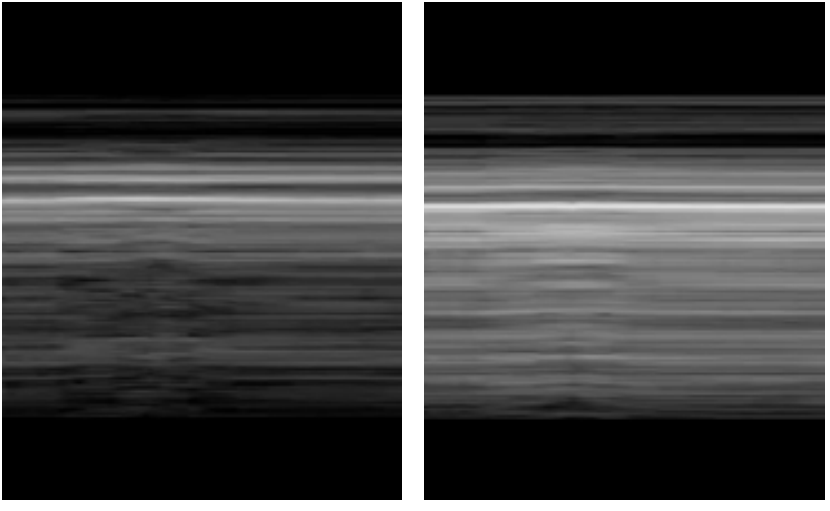}
    \caption{Absent lung sliding, barcode sign}
    \label{subfig:no-sliding2}
  \end{subfigure}
  \caption{Examples of M-mode images satisfying the pairwise relationship of belonging to the same original B-mode video and intersecting the pleural line.}
  \label{fig:mmodes-pairwise}
\end{figure*}

As outlined in \cref{subsubsec:mmodes-from-bmodes}, the brightest $50\%$ of the M-mode images within the bounds of the pleural line are utilized for pretraining. We consider any such M-mode images from the same video to be a positive pair. Qualitatively, different M-mode images produced from the same B-mode video appear very similar. Crucially, they would have the same lung sliding label, fulfilling the alignment condition outlined in~\cite{Balestriero2022}. \cref{fig:mmodes-pairwise} displays examples of positive pairs of M-modes.  We fix the pairwise relationship to focus on evaluating data augmentation transformations and the effects of pretraining under different settings, relegating an ablation study for the M-mode pairwise relationship to future work.

\subsubsection{Data Augmentation Strategy}
\label{subsubsec:data-augmentation-strategy}

A pretraining data augmentation strategy was devised to improve the invariance against inconsequential transformations and noise inherent to M-mode images. Transformations were identified that simulate natural variations present in M-mode LUS but that do not impact the patterns exhibited by absent or present lung sliding, such as speckle noise and variable depth, gain, and contrast. The following series of stochastic transformations is applied to each M-mode image prior to subjecting it to the forward pass in pretraining:

\begin{enumerate}
    \setlength\itemsep{0.2pt}
    \item With probability $0.8$, we random crop of $c \sim \mathcal{U}(0.08, 1.0)$ of the image's area, which is resized to its original dimensions. To ensure the pleural line was retained, the top of the crop was always within the upper half of the image.
    \item With probability $0.5$, horizontal flip
    \item With probability $0.5$, Gaussian blur with a horizontal kernel of $10$ pixels and $\sigma \sim \mathcal{U}(0.1, 2.0)$
    \item With probability $0.5$, random additive Gaussian noise is added to each pixel, sampled from $\mathcal{N}(\mu, \sigma^2)$, where $\mu \sim \mathcal{U}(-10, 10)$ and $\sigma \sim \mathcal{U}(0, 25)$
    \item With probability $0.5$, application of speckle noise, simulated using multiplicative Gaussian noise, sampled from $\mathcal{N}(1, \sigma^2)$, with $\sigma \sim \mathcal{U}(0, 0.1)$
    \item With probability $0.8$, brightness adjustment by \\ $c \sim \mathcal{U}(-0.4, 0.4)$
    \item With probability $0.8$, contrast adjustment by \\ $c \sim \mathcal{U}(-0.4, 0.4)$. With probability $0.5$, contrast adjustment occurs before brightness adjustment.
\end{enumerate}

\begin{figure*}
    \centering
      \begin{subfigure}{0.47\linewidth}
          \includegraphics[width=\linewidth]{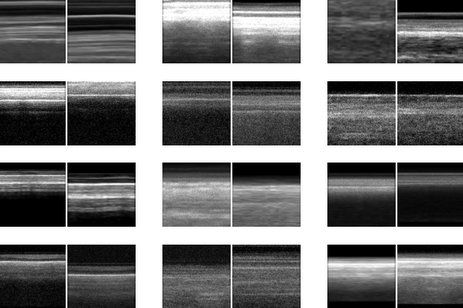}
          \caption{Custom M-mode augmentation pipeline}
          \label{subfig:mmode_augmentations}
      \end{subfigure}
      \hfill
      \begin{subfigure}{0.47\linewidth}
        \includegraphics[width=\linewidth]{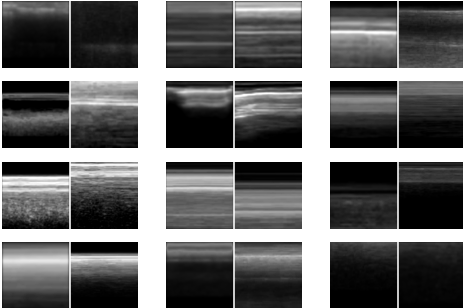}
        \caption{BYOL augmentation pipeline~\cite{Grill2020}}
        \label{subfig:byol_augmentations}
      \end{subfigure}
    \caption{A selection of M-mode positive pairs that have been subjected to the M-mode-specific data augmentation pipeline.}
    \label{fig:augmentations}
\end{figure*}

In summary, a positive pair consists of two M-mode images taken from a \SI{3}{s} segment of the same original B-mode LUS video that are then transformed via data augmentation. See \cref{fig:augmentations} for some examples of positive pairs.

\subsection{Evaluation Protocol}

\subsubsection{Lung Sliding Classification}

The downstream supervised learning task is the identification of absent lung sliding in M-mode images, which is a binary classification task. M-mode images of the upper and middle lobes of the lung are well suited for this problem, as there exist established visual patterns employed by clinicians to distinsuish between present and absent lung sliding. The \textit{seashore sign} is indicative of lung sliding~\cite{Lichtenstein2010}, whereas the \textit{barcode sign} signals absent lung sliding~\cite{Lichtenstein2005}. We consider absent lung sliding to be the positive class.

\subsubsection{Data Augmentation Pipeline Evaluation}

We first ascertain the merit of the custom M-mode LUS data augmentation pipeline for the pretext task (described in \cref{subsubsec:data-augmentation-strategy}), as opposed to the ubiquitously cited augmentation pipeline for the ``Bring Your Own Latent" (BYOL) SSL pretraining method proposed by Grill \etal~\cite{Grill2020}, which we accordingly refer to as the \textit{BYOL augmentation pipeline}. Unlike the original BYOL study, we do not apply hue changes because LUS M-mode images are greyscale. For each of the three studied pretraining methods, two CNN feature extractors are trained using $\mathcal{D}_{\text{train}}$ on the pretext task using the custom and BYOL augmentation pipelines respectively, producing a total of six feature extractors. A linear classifier is then appended and trained on the downstream task for each of the feature extractors. The best-performing augmentation pipeline across such experiments is then used for all subsequently performed pretraining trials. 

\subsubsection{Linear and Fine-tuning Evaluation}

To investigate the effect of self-supervised pretraining on the performance in the downstream task, we adopt an evaluation protocol similar to those conducted by contemporary SSL publications that evaluate on natural images. We append a single-node fully connected layer to the feature extractor with sigmoid activation.

We consider two methods to evaluate the utility of weights pretrained with SSL -- linear modelling and fine-tuning. In the linear evaluation, the weights of the pretrained model are fixed and a linear classifier is trained using the feature representations from the extractor. In the case of fine-tuning, all model weights past the first block of the network are subject to updates. In both cases, we train for $40$ epochs with a learning rate of \num{1e-4}, which is decayed by $0.03$ every epoch after the $15^{\text{th}}$ epoch. There is a heavy class imbalance in $\mathcal{D}_{\text{lab}}$ that favours present lung sliding. As a result, we oversample the minority class, taking the second, third, and fourth brightest M-modes from each absent lung sliding video. Models are trained using the binary cross entropy loss function. Data augmentation is applied to training images, with random contrast reduction by $c \sim \mathcal{U}(0, 0.3)$, random brightness adjustment by $b \sim \mathcal{U}(-0.1, 0.1)$, additive Gaussian noise sampled from $\mathcal{N}(0, 5)$, and random horizontal flip. The model weights resulting in the lowest loss on $\mathcal{D}_{\text{val}}$ were saved for evaluation.

The performance of self-supervised pretrained models are compared against two fully supervised baselines, where weights are initialized with either ImageNet-pretrained weights or random weights.

\subsubsection{Label Efficiency}

To assess the effects of SSL pretraining in its entirety, it is essential to consider its potency with respect to both performance differences and the ability to leverage otherwise unusable unlabelled datasets. Accordingly, we carry out a series of experiments that compare downstream performance under different levels of labelled data availability. We pretrain on $\mathcal{D}_{\text{train}}$ and fine-tune using progressively larger subsets of $\mathcal{D}_{\text{train}}$. Fine-tuning is repeated using ImageNet-pretrained weights and randomly initialized weights, facilitating comparisons in low-label scenarios. Lastly, we pretrain on a large dataset consisting of $\mathcal{D}_{\text{lab}}$ and $\mathcal{D}_{\text{unl}}$.

\subsubsection{Explainability}

Saliency maps (also referred to as ``heatmaps") are a type of explanation for CNN predictions that consist of the orginal image superimposed onto a colour map indicating the regions that were most contributory to the prediction. We generate saliency maps using Grad-CAM~\cite{Selvaraju2017} for selected predictions using models pretrained with SSL, initialized with ImageNet weights, and randomly initialized. The LUS expert compares the appropriateness of the saliency maps produced for pretrained and fully supervised models.

\section{Results}
\label{sec:results}

\subsection{Data Augmentation}

We investigate the effect of applying our custom M-mode data augmentation pipeline in the pretraining phase. Three models were pretrained using the examined SSL methods. We execute linear evaluation trials and compare the AUC on $\mathcal{D}_{\text{val}}$. As demonstrated in \cref{tab:augmentation_comparison}, the custom M-mode augmentation pipeline results in higher performance in linear evaluation when the pretrained models are initialized randomly. Interestingly, the BYOL augmentation pipeline exhibits a marked improvement compared to the M-mode-specific augmentations in the case of ImageNet initialization. As a result, all further experiments pretrained from scratch and from ImageNet-pretrained weights use the M-mode augmentations and BYOL augmentations respectively.

\begin{table}
    \centering
    {\small
    \begin{tabular}{lcccc}
        \toprule
        & \multicolumn{2}{c}{Random init.} & \multicolumn{2}{c}{ImageNet init.} \\
        & BYOL & M-mode & BYOL & M-mode  \\
        \midrule
        SimCLR & $0.585$ & $0.658$ & $0.864$ & $0.827$ \\
        Barlow Twins & $0.554$ & $0.578$ & $0.826$ & $0.818$ \\
        VICReg & $0.6208$ & $0.6377$ & $0.826$ & $0.798$ \\
        \bottomrule
    \end{tabular}}
    \caption{AUC on $\mathcal{D}_\text{val}$ of linear classifier trained using feature representations from various self-supervised networks.}
    \label{tab:augmentation_comparison}
\end{table}

\subsection{Comparison with Supervised Baselines}

To assess the quality of the pretrained representations, we conduct linear evaluation and fine-tuning trials using pretrained feature extractors. The results on $\mathcal{D}_{\text{test}}$ are compared with a linear classifier trained atop feature extractors initialized randomly and with ImageNet-pretrained weights. \cref{tab:test-set-results} summarizes the performance on $\mathcal{D}_{\text{test}}$. Immediately apparent is the utility of initializing all feature extractors with ImageNet-pretrained weights, including those pretrained with SSL. Among the fine-tuning trials initialized with ImageNet, self-supervised pretrained models exhibit greater test performance. The results are less clear for linear evaluation using ImageNet weight initialization, as linear models using a frozen ImageNet-pretrained feature extractor achieve the greatest test AUC and sensitivity. We additionally find that in all cases where weights are initialized randomly, self-supervised pretrained models outperform the fully supervised baselines.

\begin{table*}
    \begin{centering}
    \begin{tabular}{llcccccccc}
        \toprule
        & & \multicolumn{2}{c}{$\mathcal{D}_{\text{test}}$ AUC}
        & \multicolumn{2}{c}{$\mathcal{D}_{\text{test}}$ Specificity}
        & \multicolumn{2}{c}{$\mathcal{D}_{\text{test}}$ Sensitivity} 
        & \multicolumn{2}{c}{$\mathcal{D}_{\text{test}}$ Accuracy}\\
        Pretraining Method & Initialization & Linear & Fine-tune
        & Linear & Fine-tune
        & Linear & Fine-tune 
        & Linear & Fine-tune \\
        \midrule
        \multirow{2}{*}{{\large SimCLR}} & ImageNet & $0.742$ & $0.861$ & $\mathbf{0.764}$ & $\mathbf{0.909}$ & $0.497$ & $0.545$ & $\mathbf{0.701}$ & $\mathbf{0.823}$ \\
        & Random & $0.645$ & $\mathit{0.662}$ & $0.582$ & $0.755$ & $0.619$ & $0.476$ & $0.590$ & $0.688$ \\
        \multirow{2}{*}{{\large Barlow Twins}} & ImageNet & $0.707$ & $\mathbf{0.866}$ & $0.705$ & $0.845$ & $0.598$ & $\mathbf{0.741}$ & $0.680$ & $0.820$ \\
        & Random & $\mathit{0.646}$ & $0.634$ & $0.644$ & $0.718$ & $0.534$ & $0.460$ & $0.618$ & $0.657$ \\
        \multirow{2}{*}{{\large VICReg}} & ImageNet & $0.738$ & $0.834$ & $0.661$ & $0.84$ & $0.703$ & $0.656$ & $0.671$ & $0.797$ \\
        & Random & $0.609$ & $0.619$ & $\mathit{0.926}$ & $\mathit{0.860}$ & $0.286$ & $0.318$ & $\mathit{0.774}$ & $\mathit{0.731}$ \\
        \multirow{2}{*}{{\large None}} & ImageNet & $\mathbf{0.768}$ & $0.854$ & $0.638$ & $0.834$ & $\mathbf{0.751}$ & $0.714$ & $0.665$ & $0.805$ \\
        & Random & $0.500$ & $0.585$ & $0.000$ & $0.563$ & $\mathit{1.000}$ & $\mathit{0.540}$ & $0.237$ & $0.558$ \\
        \bottomrule
    \end{tabular} 
    \caption{Downstream performance on $\mathcal{D}_\text{test}$, trained $\mathcal{D}_\text{train}$. \textit{Typescript} entries correspond to the best performance when using randomly initialized weights and \textbf{boldface} entries identify the best performance for trials initialized with ImageNet-pretrained weights.}
    \label{tab:test-set-results}
    \end{centering}
\end{table*}

\subsection{Label Efficiency}

\begin{figure}
    \centering
    \includegraphics[width=\linewidth]{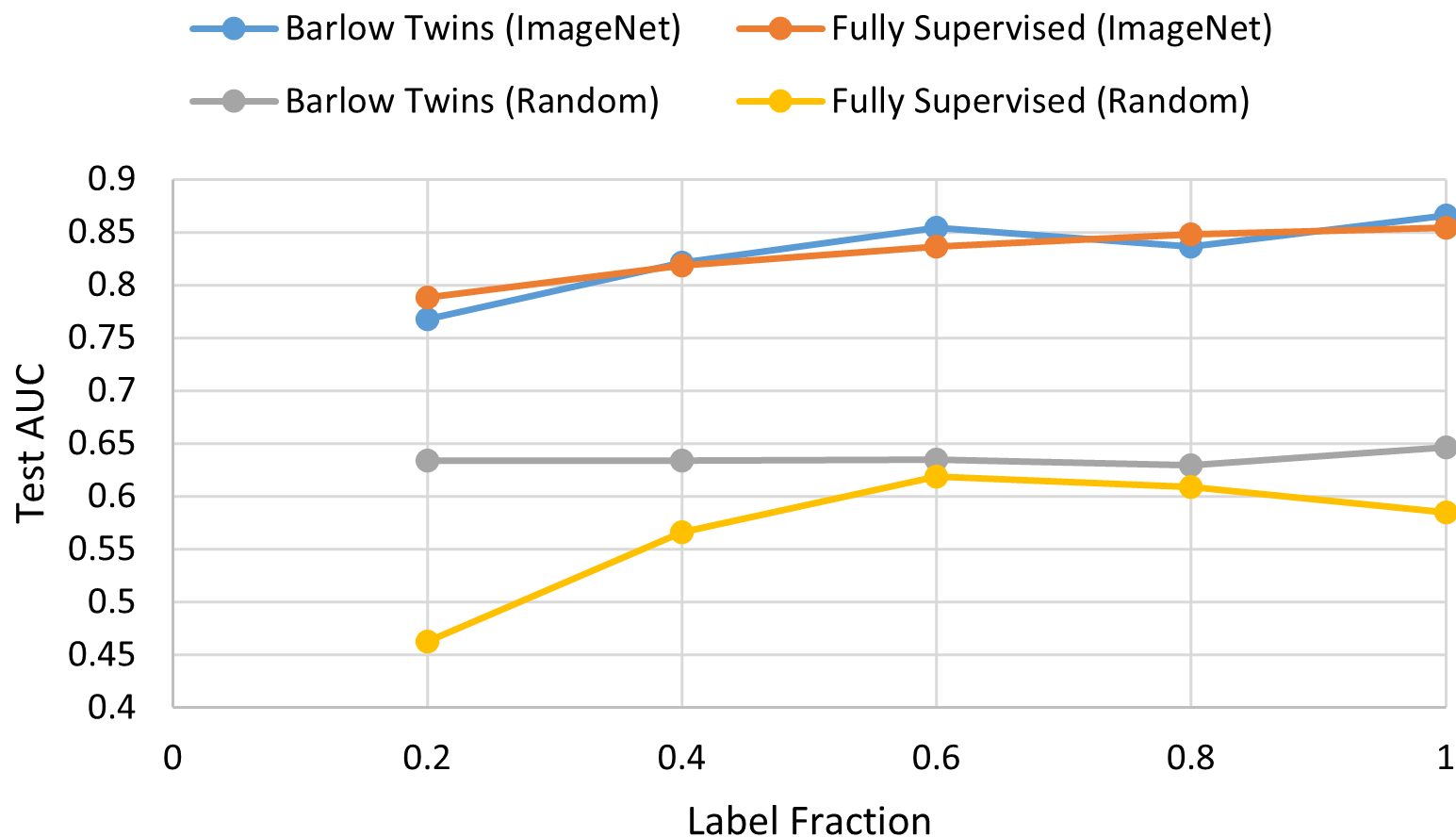}
    \caption{AUC on $\mathcal{D}_{\text{test}}$ for fine-tuned models initialized from weights pretrained using Barlow Twins and fully supervised models initialized randomly or with ImageNet-pretrained weights.}
    \label{fig:train-prop-experiment}
\end{figure}

One of the major benefits of SSL is its ability to leverage unlabelled examples. To measure the effect of varying proportions of labelled data, we drop fractions of the labels in $\mathcal{D}_{\text{train}}$ and conduct fine-tuning using networks pretrained using Barlow Twins on all examples in $\mathcal{D}_{\text{train}}$. \cref{fig:train-prop-experiment} details the results. The feature extractor trained from scratch benefitted from self-supervised pretraining in the low-label setting. However, it appears that this benefit is greatly diminished when the pretrained feature extractor and fully supervised feature extractor are both initialized with weights pretrained on ImageNet.

To further elucidate the benefit of pretraining with greater volumes of unlabelled data, we pretrain on both $\mathcal{D}_{\text{train}}$ and $\mathcal{D}_{\text{unl}}$ using Barlow Twins and compare the performance to pretraining with $\mathcal{D}_{\text{train}}$ alone. As shown in \cref{tab:unlabelled-comparison}, more unlabelled data greatly improves the test performance of the feature extractors pretrained using random initialization, while performance drops for those initialized with ImageNet weights.

\begin{table}
    \centering
    \begin{tabular}{ccc}
        \toprule
        Initialization & Pretraining data & $\mathcal{D}_{\text{test}}$ AUC \\
        \midrule
        \multirow{2}{*}{Random} & $\mathcal{D}_{\text{train}}$ & $0.634$ \\
        & $\mathcal{D}_{\text{train}} + \mathcal{D}_{\text{unl}}$ & $0.799$ \\
        \multirow{2}{*}{ImageNet} &  $\mathcal{D}_{\text{train}}$ & $0.866$ \\
        & $\mathcal{D}_{\text{train}} + \mathcal{D}_{\text{unl}}$ & $0.838$ \\
        \bottomrule
    \end{tabular}
    \caption{Downstream classification performance of models pretrained using Barlow Net using labelled and unlabelled datasets.}
    \label{tab:unlabelled-comparison}
\end{table}

\subsection{Evaluation on External Datasets}

We evaluate all pretrained and fully supervised fine-tuned models initialized with ImageNet-pretrained weights on $\mathcal{D}_\text{ext1}$, $\mathcal{D}_\text{ext2}$, and $\mathcal{D}_\text{ext3}$. The results, presented in \cref{tab:external_performance}, do not indicate that the self-supervised models pretrained on $\mathcal{D}_\text{train}$ outperform the fully supervised baseline when evaluated on datasets originating from other centres. However, with the exception of sensitivity, we find that performance on external datasets distinctly increases when models are pretrained on $\mathcal{D}_\text{train}$ and $\mathcal{D}_\text{unl}$ combined, highlighting a potential benefit of leveraging unlabelled data when labels are partially available.

\begin{table*}[]
    \centering
    {\small
    \begin{tabular}{lccccc}
        \toprule
         & Pretraining dataset(s) & AUC & Specificity & Sensitivity & Accuracy \\
         \midrule
         \multirow{2}{*}{SimCLR} & $\mathcal{D}_{\text{train}}$ & $0.765\,[0.063]$ & $0.765\, [0.043]$  & $0.532 \,[0.042]$  & $0.729\, [0.037]$  \\
         
         & $\mathcal{D}_{\text{train}}$ + $\mathcal{D}_{\text{unl}}$ & $0.778\, [0.053]$ & $\mathbf{0.862 \,[0.046]}$ & $0.429 \,[0.103]$ & $\mathbf{0.834 \,[0.056]}$ \\

         \multirow{2}{*}{Barlow Twins} & $\mathcal{D}_{\text{train}}$ & $0.802 \,[0.071]$ & $0.728 \,[0.043]$ & $0.727 \,[0.138]$ & $0.731\, [0.048]$ \\

         & $\mathcal{D}_{\text{train}}$ + $\mathcal{D}_{\text{unl}}$ & $\mathbf{0.833 \,[0.059]}$ & $0.761 \,[0.042]$ & $0.723 \,[0.076]$ & $0.756\, [0.036]$ \\
         
         \multirow{2}{*}{VICReg} & $\mathcal{D}_{\text{train}}$ & $0.807\, [0.048]$ & $0.727 \,[0.073]$ & $\mathbf{0.785 \,[0.021]}$  & $0.740\, [0.059]$ \\

         & $\mathcal{D}_{\text{train}}$ + $\mathcal{D}_{\text{unl}}$ & $0.817 \,[0.053]$ & $0.705 \,[0.076]$ & $0.760\, [0.114]$ & $0.720\, [0.054]$ \\
         
         Fully Supervised & - & $0.804 \,[0.076]$ & $0.682\, [0.097]$ & $0.761 \,[0.075]$ & $0.699 \,[0.071]$ \\
         \bottomrule
    \end{tabular}}
    \caption{Mean [std] performance across the three external datasets for pretrained and fully supervised models. Each was initialized with ImageNet weights and (pre)trained using $\mathcal{D}_{\text{train}}$. Models pretrained using $\mathcal{D}_{\text{train}}$ and $\mathcal{D}_{\text{unl}}$ outperformed those pretrained using $\mathcal{D}_{\text{train}}$ alone.}
    \label{tab:external_performance}
\end{table*}

\subsection{Explainability}

\begin{figure}
  \centering
  \begin{subfigure}{0.48\linewidth}
    \includegraphics[width=\linewidth]{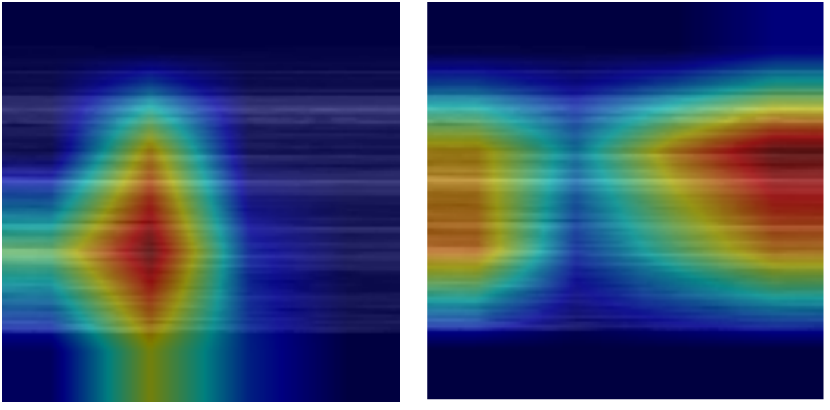}
    \caption{False positive for both models.}
    \label{fig:hm1}
  \end{subfigure}
  \hfill
  \begin{subfigure}{0.48\linewidth}
    \includegraphics[width=\linewidth]{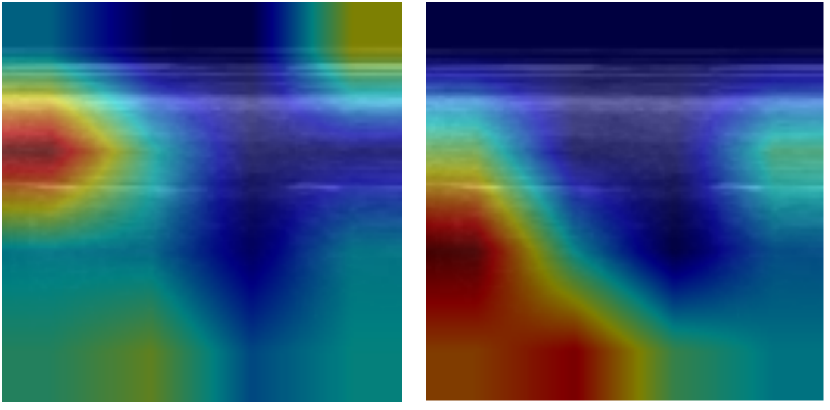}
    \caption{True negative for both models.}
    \label{fig:hm2}
  \end{subfigure}
  \hfill
  \begin{subfigure}{0.48\linewidth}
    \includegraphics[width=\linewidth]{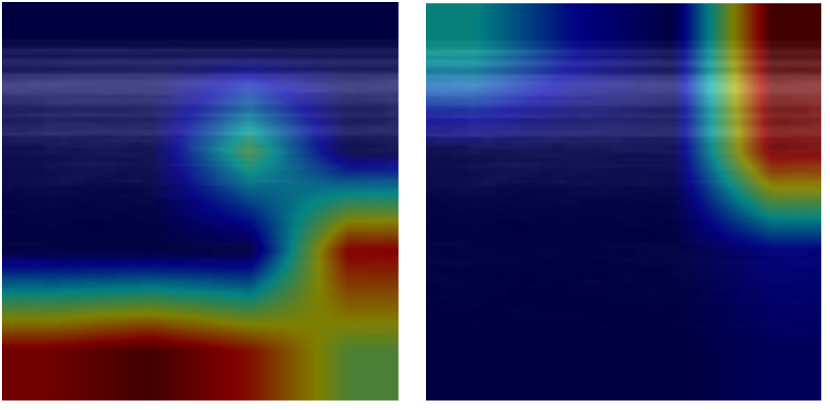}
    \caption{False negative for the fully supervised model; true positive for the pretrained model.}
    \label{fig:hm3}
  \end{subfigure}
  \hfill
  \begin{subfigure}{0.48\linewidth}
    \includegraphics[width=\linewidth]{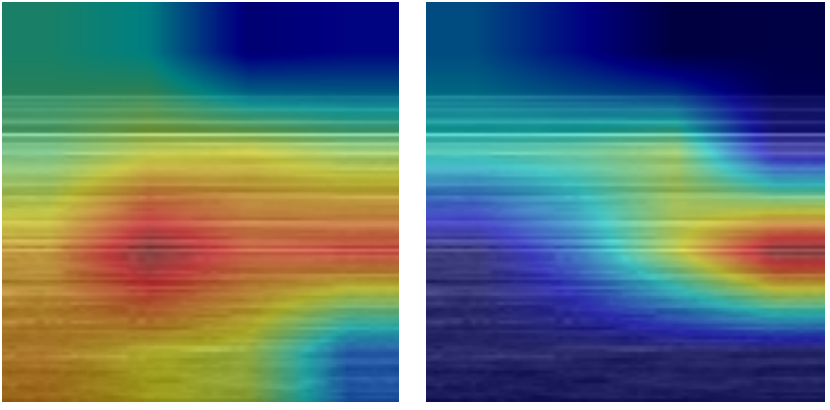}
    \caption{True positive for both models.}
     \vspace{18pt}
    \label{fig:hm4}
  \end{subfigure}
  \caption{A selection of saliency maps for predictions produced by both a fully supervised model (left in each subfigure) and a model pretrained with Barlow Twins. }
  \label{fig:gradcam}
\end{figure}

To explore the patterns learned by the fine-tuned models, we select the pretrained network with the greatest $\mathcal{D}_{\text{test}}$ AUC (Barlow Twins) and produce saliency maps for $4$ M-mode images in $\mathcal{D}_{\text{test}}$, using the pretrained network and a fully supervised network initialized with ImageNet-pretrained weights (see \cref{fig:gradcam}). The LUS expert reviewed the saliency maps without knowing which method was used to produce them and rated the appropriateness of the heatmaps using a $4$-point Likert scale from $0$ to $3$, based on whether the highlighted regions correspond to the areas of clinical interest. For instance, despite the false positive prediction by the self-supervised model shown in \cref{fig:hm1}, the far field (bottom of the image) is dark, and the saliency map indicates activation by the poignant straight lines in the near field, producing a prediction of absent lung sliding. The saliency maps generated for the pretrained model and fully supervised model scored averages of $2.5$ and $1.5$ respectively.

\section{Discussion}
\label{sec:discussion}

The results in this study characterize the utility of self-supervised pretraining for M-mode LUS examinations. First, it is clear that fine-tuning a self-supervised feature extractor provides improved initial feature representations compared to randomly initialized weights. The gap in performance on unseen examples strikingly narrows when pretrained models and their fully supervised counterparts are initialized using the omnipresent ImageNet-pretrained weights. These results indicate that, while there may be performance gains when fine-tuning SSL-pretrained models, the state-of-the-art constrastive and nonconstrative methods alone are not sufficient to reap substantial gains when trained on the labelled dataset and initialized with ImageNet-pretrained weights. However, when training from scratch, practitioners may benefit from pretraining.

A major insight derived from the results is that SSL-pretrained models improve performance on datasets from external centres for the task of lung sliding classification when using large volumes of unlabelled data that do not appear in the training set. Generalizability is a paramount concern for any organization deploying machine learning systems in healthcare settings. This finding is consistent with the results of studies showing that SSL pretraining improves performance on external data in other medical imaging domains, such as chest X-ray classification~\cite{Azizi2021, Tiu2022}, dermatologic image classification~\cite{Azizi2021}, and pathology slide classification~\cite{Zheng2022}. Practitioners seeking to promote generalizability are therefore encouraged to employ self-supervised pretraining using any available unlabelled data.

Another noteworthy finding is that ImageNet-initialized feature extractors pretrained with the BYOL augmentation pipeline yield better performance for the downstream classification task. Again, the duality between ImageNet-pretrained and randomly initialized feature extractors manifests itself, as the M-mode augmentation pipeline produces better feature representations when pretrained models were initialized randomly. We conjecture that the BYOL augmentation pipeline's efficacy is related to the fact that the pretrained models were initialized with ImageNet-pretrained weights, since the original BYOL paper (and subsequent SSL publications, such as~\cite{Zbontar2021, Bardes2022}) focused on improving downstream tasks with ImageNet~\cite{Grill2020}.

The present work has notable limitations. The training set consisted of LUS examinations collected within a single healthcare institution, thereby limiting heterogeneity with respect to sources of variance such as device manufacturer, practitioner skill sets, and patient populations. Secondly, SimCLR performs best when larger batch sizes are employed during pretraining~\cite{Chen2020}; however, due to material limitations, we utilize a considerably small batch size. In keeping with the authors' findings, we train for a large number of epochs to mitigate the impact of a small batch size. Lastly, this study did not meet or exceed the performance metrics reported in the most recent publication regarding automatic lung sliding classification~\cite{VanBerlo2022}. However, unlike~\cite{VanBerlo2022}, we employ different datasets, we use standard binary cross entropy loss, we do not apply any techniques to mitigate overfitting, and we do not add fully connected layers between the feature extractor and the output layer. Rather than aiming to maximize the performance of the classifier, our focus is to study the effect of different SSL strategies and data augmentation distributions on the quality of representations.

There are multiple avenues for future work. First, a subsequent investigation could undertake a comprehensive inquiry into the data augmentation transformations for joint-embedding SSL methods applied to M-mode LUS. secondly, the lack of consensus regarding the augmentation pipeline motivates future work concerning the underlying reasons and possible discovery of novel M-mode ultrasound data transformations. Lastly, further research could explore alternative pretext tasks. In this study, we propose a pairwise relationship for M-mode images consisting of images that were taken from the same B-mode video, adopting common contrastive and non-contrastive learning pretext tasks; however, a novel pretext task could be formulated to better suit the M-mode ultrasound domain.

\section{Conclusion}

A selection of contemporary contrastive and non-contrastive SSL pretraining methods were investigated for LUS M-mode data, using a pairwise relationship specific to M-mode LUS. When evaluated on the downstream binary classification task of absent lung sliding detection, fine-tuned feature extractors initialized with self-supervised pretrained weights generally exhibited greater performance than fully supervised counterparts. Pretraining with larger unlabelled datasets resulted in improved metrics on evaluation datasets from external institutions. The results spur multiple directions for future work, such as the refinement of M-mode ultrasound data augmentation pipelines for SSL and the evaluation of alternative or novel pretext tasks.

{\small
\bibliographystyle{ieee_fullname}
\bibliography{citations}
}

\end{document}